\documentclass[amsmath, amssymb, aps, pra, twocolumn,floatfix,nofootinbib]{revtex4}
\usepackage[utf8]{inputenc}
\usepackage{multirow}
\usepackage[urlcolor=blue, colorlinks=true,citecolor=blue]{hyperref}
\usepackage{enumitem}

\usepackage{amsmath}
\usepackage{amsmath, amssymb}
\usepackage{textcomp, gensymb}
\usepackage{tabularx}

\usepackage{wrapfig}
\usepackage{graphicx}
\usepackage{float}
\usepackage{subfigure}
\usepackage{amssymb}
\usepackage{bbold}
\usepackage{color}
\usepackage{ulem}
\usepackage{algorithm}
\usepackage[version=4]{mhchem}
\usepackage{algpseudocode} 

\begin{document}

\title{Stabilization of industrial processes\\with time series machine learning}

\author{Matvei Anoshin}
\author{Olga Tsurkan}
\author{Vadim Lopatkin}
\author{Leonid Fedichkin}
\affiliation{L.D. Landau Dept. of Theoretical Physics, Moscow Institute of Physics and Technology, Institutsky Per. 9, Dolgoprudny, Moscow Region, 141701 Russia}


\begin{abstract}
The stabilization of time series processes is a crucial problem that is ubiquitous in various industrial fields. The application of machine learning to its solution can have a decisive impact, improving both the quality of the resulting stabilization with less computational resources required. In this work, we present a simple pipeline consisting of two neural networks: the oracle predictor and the optimizer, proposing a substitution of the point-wise values optimization to the problem of the neural network training, which successfully improves stability in terms of the temperature control by about $3$ times compared to ordinary solvers.                                                       
\end{abstract}

\maketitle

\section*{Introduction}

An application of machine learning to the industrial processes stabilization is an open problem which promises a huge potential benefit to the such critical industries as metals and energy development if solved.

Classical optimization methods, such as finite-horizon markov decision processes \cite{Mundhenk_Goldsmith_Lusena_Allender_2000}, non-linear programming reformulation of control \cite{Trierweiler_2014} and point-wise optimization \cite{Niegodajew_Marek_Elsner_Kowalczyk_2020} are frequently employed in order to achieve better stability of time series process, successfully improving production quality, minimizing expenses and manufacturing devices deficiency with near-future planing or real-time optimization.  

Machine learning, known for its prominent results in solution of enterprise problems \cite{Cruz_Villalonga_Castaño_Rivas_Haber_2023}, became widely applied to the time series prediction and generation after recent advances in such fields as natural language processing, due to the similarity aforementioned tasks in their time dependent recurrent nature \cite{Dalal_Verma_Chahar_2024}. Thus, contemporary time series modeling is performed with long short-term memory (LSTM) models \cite{Hochreiter_Schmidhuber_1997} and Transformers \cite{vaswani2023attentionneed}, incorporating different attention strategies. 

Currently, state-of-the-art approaches to ML-driven optimization include an application of reinforcement learning, but for time series problems, the usual focus stays on approximation of the industrial process as a dynamic system on the basis of recurrent neural network (RNN), with such methods as recurrent stabilization control \cite{Kulawski_Brdyś_2000, gu2021recurrentneuralnetworkcontrollers}. After dynamical model is obtained, any classical approaches such as proportional–integral–derivative (PID) \cite{pid} controller may be applied, resulting in the stabilization of the process.

In the following work we present simple, yet powerful solution to the problem of ML-driven optimization of the ongoing physical process, showing its vices over ordinary point-wise optimization through different solvers application and PID regulation via substitution of controllable values optimization problem, to the problem of the weights approximation in the optimizer neural network. This substitution yields the better results in both terms of the work-time efficiency and optimization accuracy.

\section*{Results}

\subsection{Classical Optimization Methodology}

Let's consider an industrial process as multi-modal time series data with inner correlations $X\left(t\right) = \left(x^1\left(t\right), x^2\left(t\right), ..., x^n\left(t\right)\right)$ recorded with a given discretization time step $\Delta t$ and containing $K$ controlled, $M$ controlling and $1$ target time-series $\left(K+M+1=n\right)$.
For the sake of generalization, controlled features may be denoted as non-adjustable $x^i_\mathrm{non-adjustable}\left(t\right)$, controlling as $x^i_\mathrm{adjustable}\left(t\right)$ and target as $x_\mathrm{target}\left(t\right)$.

\begin{figure}[!h]
    \centering
    \includegraphics[width=\linewidth]{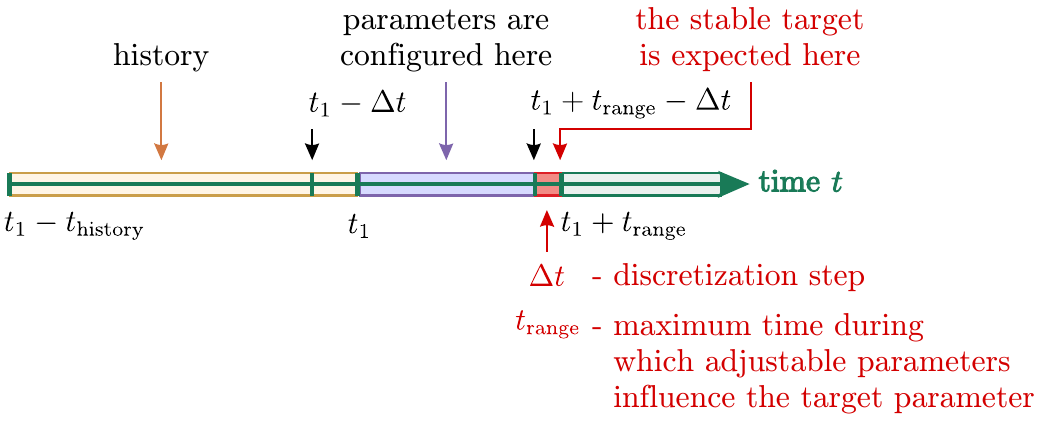}
    \caption{A time axis illustrating scales of $t_{\mathrm{range}}$ and $\Delta t$. It is assumed that  optimization policy on $t_{\mathrm{range}} - \Delta t$ mostly determines target value.}
    \label{fig:optim-interval}
\end{figure}

An optimization problem statement is prescribed as follows: 

\begin{itemize}
    \item Dataset contains a historical data of the response of $x_\mathrm{target}\left(t\right)$ and $x^i_\mathrm{non-adjustable}\left(t\right)$ on change of $x^i_\mathrm{adjustable}\left(t\right)$ parameters from $t_{\mathrm{min}}$ to $t_1$
    \item The final goal of the optimization is to devise a method, which allows to stabilize $x_\mathrm{target}\left(t\right)$ from $t \geq t_1 + t_\mathrm{range}$ by changing $x^i_\mathrm{adjustable}\left(t\right)$ values on $[t_1, t_1 + t_{\mathrm{range}}]$ as it is illustrated on Fig. \ref{fig:optim-interval}. That means what for any arbitrary chosen interval $[t_2, t_3] \in [t_1 + t_{\mathrm{range}}, +\infty]$ 
    $$\sum_{t=t_2}^{t=t_3} |x_{\mathrm{target}}\left(t\right) - x_{\mathrm{ideal}}| \rightarrow \mathrm{min}$$
\end{itemize}

The classical optimization paradigm, which is specified in Alg. \ref{algorithm:classic-pointwise}, thrives to optimize values for $x_{\mathrm{target}}\left(t\right)$ point-wise (w.r.t to $x^i_\mathrm{adjustable}\left(t\right)$), assuming that local minimum on the $i$-th step will result in the minimum on every arbitrary chosen interval. This approach, generally speaking, an exponentially hard on each $i$-th step, thus both resources and time consuming, what is especially inconvenient in case of the industrial process optimization, as industrial data usual span a vast amount of sensors with high discretization step. 

\begin{algorithm}[H]
    \caption{Classical point-wise target control}
    \begin{algorithmic}
        \State 1. On the $i$-th step of the optimization, $M\cdot t_{\mathrm{range}} / \Delta t$ values $x^j_\mathrm{adjustable}\left(t\right)|_{t=t_i}^{t=t_i - \Delta t +t_{\mathrm{range}}}$ are optimized with the cost function $$\mathrm{Cost}(x_{t_i}) = |x_{\mathrm{target}}\left(t_i + t_{\mathrm{range}}\right) - x_\mathrm{ideal}|$$
        
        \State 2. Then, only  $x^j_\mathrm{adjustable}\left(t\right)|_{t=t_i}$ are set

        \State 3. On the $i + 1$-th step of the optimization, steps 1-2 are applied to $x^j_\mathrm{adjustable}\left(t\right)|_{t=t_i + \Delta t}^{t=t_i +t_{\mathrm{range}}}$ 
    \end{algorithmic}
    \label{algorithm:classic-pointwise}
\end{algorithm}

\subsection{ML-driven Optimization Methodology}

To solve the inconveniency of the point-wise classical optimization and reduce a necessity of the iterative solution of the NP-hard problem for any $t_\mathrm{range}$ data snippet requiring optimization, a ML-driven approach for target control is proposed. 

The key idea of the following section is that training a machine learning model $M_{\mathrm{optim}}\left(x, w\right): X\left(t\right)|_t^{t-\Delta t+t_\mathrm{range}}\rightarrow x^j_\mathrm{adjustable}\left(t\right)|_{t}^{t - \Delta t +t_{\mathrm{range}}}$ to mimic the best optimization policy of the industrial process target control, substitutes the point-wise optimization of $x^i_\mathrm{adjustable}$ values to the problem of finding the best approximation of the model's weights $w$. The training of $M_{\mathrm{optim}}\left(x, w\right)$ involves the Oracle function $f$, capable of predicting the resulting $x_{\mathrm{target}}\left(t_i + t_{\mathrm{range}}\right)$ on the basis of $X\left(t\right)|_{t_i}^{t_i+t_\mathrm{range} - \Delta t}$ and required for weights tuning through a loss-function backpropagation in Stochastic Gradient Descent (SGD). 

\begin{algorithm}[H]
    \caption{ML-driven target control training}
    \begin{algorithmic}
         \State 1. On the $i$-th step of the algorithm, optimizer $M_{\mathrm{optim}}\left(x, w\right)$ maps $\left(K + 1\right)\cdot t_{\mathrm{range}} / \Delta t$ values of $\left(x^j_\mathrm{non-adjustable},\; x_\mathrm{target}\right)$ into the best optimization policy values $\hat{x}^j_\mathrm{adjustable}$ 
         
        \State 2. Then, the Oracle function $f$ predicts the value of $x_{\mathrm{target}}\left(t_i +  t_{\mathrm{range}}\right) =f\left[\left(\hat{x}^j_\mathrm{adjustable},\; x^j_\mathrm{non-adjustable},\; x_\mathrm{target}\right)\right]$
        
        \State 3. Loss function 
        $\mathrm{Loss}\left(x_{t_i}\right) = \left|x_{\mathrm{target}}\left(t_i +  t_{\mathrm{range}}\right)  - x_{\mathrm{ideal}}\right|$ is used to tune weights $w$ of $M_{\mathrm{optim}}\left(x, w\right)$ through backpropagation: $w^{(i+1)} := w^i - \alpha \frac{\partial Loss}{\partial w^i}$ 

        \State 4. On the $i + 1$-th step of the training, steps 1-3 are repeated to $x^j_\mathrm{adjustable}\left(t\right)|_{t=t_i + \Delta t}^{t=t_i +t_{\mathrm{range}}}$ and so on
    \end{algorithmic}
    \label{algorithm:ML-driven-training}
\end{algorithm}

In Alg. \ref{algorithm:ML-driven-training}, the training of the $M_{\mathrm{optim}}\left(x, w\right)$ is iteratively explained for each of $\left \lceil{(t_1 - t_\mathrm{min})/t_{\mathrm{range}}}\right \rceil$ windows of historical data. After training is performed, same $M_{\mathrm{optim}}\left(x, w\right)$ may be applied to the stabilization of some industrial process on any unseen data, which we propose to do autoregressively (Section \ref{Results}). Key advantages of the novel methodology include its time-resources efficiency, which is shown in Section \ref{Results} and scaling possibility, as the Oracle expressivity defines quality of the optimization policy in its prediction of the system feedback. 

\subsection{Dataset}

\begingroup
\newlength{\xfigwd}
\setlength{\xfigwd}{2\columnwidth}
\begin{figure*}
    \centering
    \includegraphics[width=1.05\xfigwd]{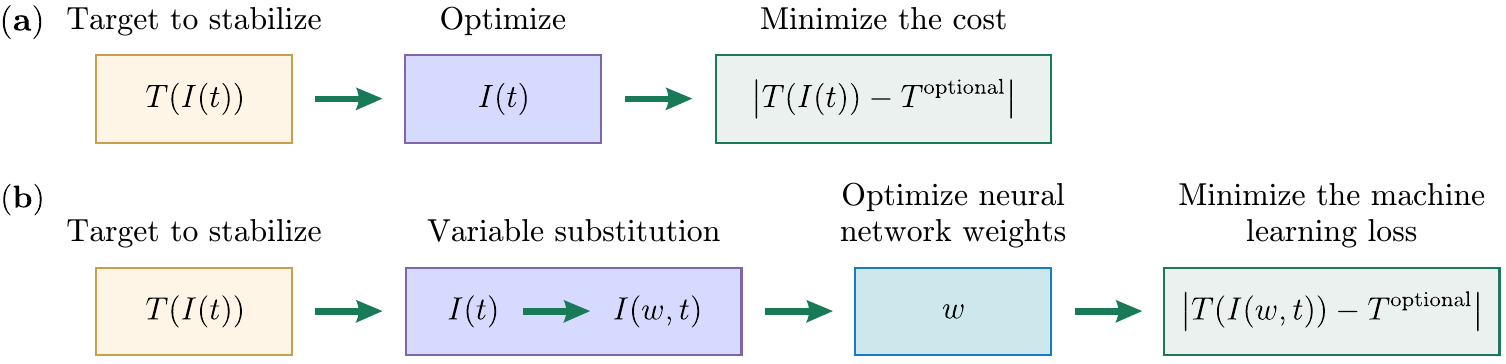}
    \caption{An explicit comparison of a) point-wise optimization to the b) ml-driven optimization, based on the proposed variables to weights optimization substitution (Alg. \ref{algorithm:ML-driven-training})}
    \label{fig:LSTMpred}
\end{figure*}
\endgroup

To comprehensively test proposed methodology, a synthetic dataset based on a thermal conductivity problem is assembled. A key idea is to simulate a practically relevant problem of room temperature $T\left(t\right)$ control with a change of the heater current $I\left(t\right)$ based on the real experimental measurements. For that purpose, air properties (absolute humidity $\mathrm{AH}\left(t\right)$ and pressure $p\left(t\right)$) in the room and environmental temperature data $T_0$ is collected. 

The differential equation describing the interaction between room and external environment presented as Newton's law of cooling with constant of losses $k$ \cite{Davidzon_2012} with an addition of the Joules heating term $I\left(t\right)^2 \cdot R$ $\left(R = 10 \ \Omega\right)$:

\begin{equation} \label{eq:thermal}
\frac{\partial T}{\partial t} = \frac{I\left(t\right)^2 \cdot R}{C_p} - \frac{k}{C_p} \left(T\left(t\right) - T_0\left(t\right)\right)
\end{equation}

With $C_p = C_0 + \alpha \cdot \text{SH}\left(t\right)$, $\alpha = 1.82 \ \frac{\mathrm{kJ}}{\mathrm{kg} \cdot ^\mathrm{\circ}\mathrm{C}}$  --- an impact factor of the special humidity (which can be obtained from $\mathrm{AH}\left(t\right)$ and $p\left(t\right)$) on the heat capacity of moist air, $C_0 = 1.005 \  \frac{\mathrm{kJ}}{\mathrm{kg} \cdot ^\mathrm{\circ}\mathrm{C}}$ - constant corresponding to the normal air heat capacity \cite{Wong_Embleton_1984}.

A historical current data $I\left(t\right)$ is modeled as:
$$ I\left(t\right) = C_0 \sin\left(\Omega_0 t\right) +  \sum_{i=1}^2 C_i  \exp\left(-\omega_i t\right) \left|\sin(\Omega_i t)\right| + \mathcal{N}\left(0, 0.2\right) $$

With randomly sampled $\left\{C_i,\; \omega_i, \; \Omega_i\right\}$ such a way, to satisfy characteristic time of the exterior temperature change $t\propto 10 
\ \mathrm{hrs}$  and characteristic amplitude of the current $I \propto \mathrm{A}$.

\begin{figure}[!h]
    \centering
    \includegraphics[width=\linewidth]{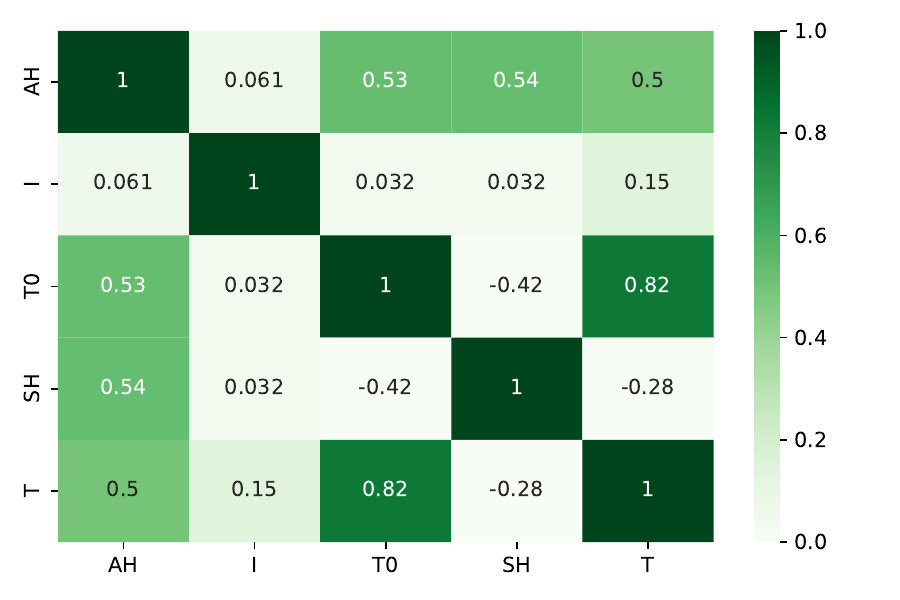}
    \caption{Correlations between real non-adjustable (AH, SH, $T_0$) and synthetic features ($T$, $I$) in the final dataset.}
    \label{fig:data}
\end{figure}

On that foundation, a finite-difference Scipy \cite{2020SciPy-NMeth} solution to the presented mathematical model of thermal conductivity is performed and stored as the dataset: $x^i_\mathrm{non-adjustable}\left(t\right) = \left(\mathrm{AH}\left(t\right),\; \mathrm{SH}\left(t\right),\; T_0\left(t\right)\right)$, $x_\mathrm{adjustable}\left(t\right) = I\left(t\right)$ and $x_\mathrm{target}\left(t\right) = T\left(t\right)$ (Fig. \ref{fig:data}).

\subsection{Training}

\begingroup
\setlength{\xfigwd}{2\columnwidth}
\begin{figure*}
    \centering
    \includegraphics[width=1.05\xfigwd]{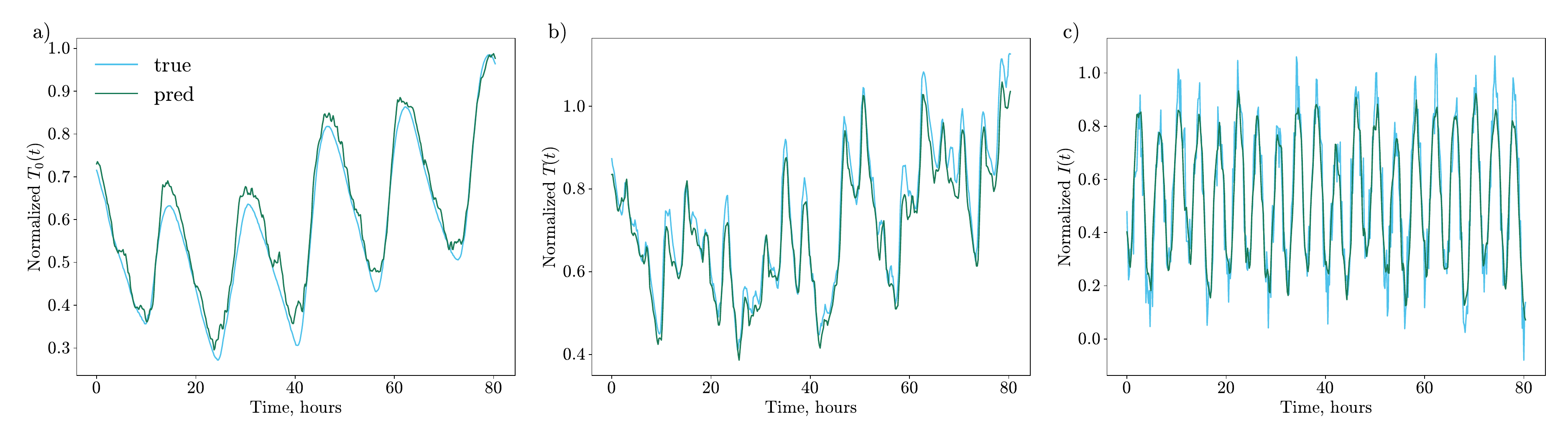}
    \caption{Prediction of the LSTM (Oracle) model for normalized values of a) exterior temperature $T_0\left(t\right)$, b) inside temperature $T\left(t\right)$ and c) heater current $I\left(t\right)$.}
    \label{fig:LSTMpred}
\end{figure*}
\endgroup

The training process covered two main phases, such as the training of the physical predictor model (the Oracle $f$) and the optimizer model respectively. For models evaluation, the synthetic dataset $X\left(t\right)|_{t_1}^{t_{2764}}$ with length of $2764$ points was dived into train and test subsets (with ratio = $0.2$) and then separated into windows such as $X_{train_t} = X\left(t\right)_t^{t + 23\Delta t}$ and $y_{train_t} =  X\left(t\right)_{t + 24\Delta t}^{t + 28\Delta t}$ with shift between each of windows $s = 1$. Each model was implemented via PyTorch \cite{PyTorch} on Python.

\subsubsection{Training of the Predictor}

At first, the Oracle implemented as enhanced LSTM model consisting of a LSTM cell (input$\_$size = $24$, hidden$\_$size = $143$, num$\_$layers = $2$) and two linear layers (n$\_$layers$_1$ = $804$, n$\_$layers$_2$ = $5$) with ReLU activation functions was trained for $50$ epochs on the training slice of the dataset to predict all 5 features in the dataset. This is performed through Adam backpropagation of $\left|\left|y_{train_i} - f\left(X_{train_t}\right)\right|\right|_2^2$.
 
Therefore, for $i$-th input vector of historical data with length of $t_\mathrm{range} = 24 \Delta t$ steps in the past ${X\left(t\right)}_{\tau-23\Delta t}^{\tau}$ resulting prediction would be an approximation for next $5 \Delta t$ steps ${X\left(t\right)}_{\tau + \Delta t}^{\tau + 5\Delta t}$ obtained as:

$${X\left(t\right)}_{\tau + \Delta t}^{\tau + 5\Delta t} = \mathrm{ReLU}\left(W_2\cdot\mathrm{ReLU}(W_1 \cdot \mathrm{LSTM}\left({X\left(t\right)}_{\tau-23\Delta t}^{\tau}\right)\right) $$ 

$$W_1 \in \mathbb{R}^{\mathrm{n}\_\mathrm{layers}_1\times \mathrm{hidden}\_\mathrm{size}},\; W_2 \in
\mathbb{R}^{\mathrm{n}\_\mathrm{layers}_2 \times \mathrm{n}\_\mathrm{layers}_1}$$  

This results in the fair approximation of ongoing physical process on the test slice of the data (Fig. \ref{fig:LSTMpred}), as model predicts features with an accuracy of $0.002 \div 0.01$ in terms of MSE.

\subsubsection{Training of the Optimizer}

At second, the optimizer model $M_{\mathrm{optim}}\left(x, w\right)$ presenting a liner regression with ReLU activation  meant to predict a correct sequence of currents $${I\left(t\right)}_{\tau + 1\Delta t}^{\tau + 5\Delta t} = M_{\mathrm{optim}}\left(X\left(t\right)|_{\tau - 23\Delta t}^{\tau},\;w\right)  = \mathrm{ReLU}\left(X\left(t\right) \cdot w\right)$$ 
$$w \in \mathbb{R}^{\mathrm{n}\_\mathrm{layers}_2\times \mathrm{hidden}\_\mathrm{size}}$$ 
consequentially minimizing the L1 difference between features temperatures ${T\left(t\right)}_{\tau + \Delta t}^{\tau + 5\Delta t}$ and ideal value $T_{ideal} = 23^\circ$. Described structure was trained for $50$ epochs on the train slice as Alg. \ref{algorithm:ML-driven-training} suggests.

\subsection{Results}\label{Results}

To perform the final stabilization on the test slice of the data, an autoregressive approach to the optimization is utilized. In order to provide its a brief explanation, let's denote $t_0$ as the starting point in the test sequence. Then, for $I_{24}$, an optimization policy established as $M_{\mathrm{optim}}\left(X\left(t\right)|_{t_0}^{t_{0} + 23\Delta t},\; w\right) \bigl[ 0\bigr] $ ($v\bigl[ i\bigr]$ - corresponds to the $i$-th component of vector $v$). For $I_{25}$, optimizer is applied to the shifted on $s=1$ interval $\hat{X}\left(t\right)|_{t_1}^{t_{24}} = X\left(t\right)|_{t_1}^{t_{23}} + f\left(X\left(t\right)|_{t_0}^{t_{23}}\right)$, with $f$ - the Oracle function. Therefore, we explicitly include the change of the physical environment and its response due to the intervention with new optimization policy.

Autoreggressive optimization shows promising results Fig. \ref{fig:OptimRes}a), successfully stabilizing the room temperature around $23 \pm 1.07 \ ^\circ C$.  

As benchmarks of the proposed methodology, two classical approaches are suggested. The first one is Proportional-Integral-Derivative (PID) controller algorithm, famously known as standard approach in such industrial areas, as thermal control. To fairly evaluate its performance, the numerical solution of Eq. \ref{eq:thermal} with account of PID regulated currents is performed (with other non-adjustable values taken from the test slice $X\left(t\right)$'s) Fig. \ref{fig:OptimRes}b).

The second one is more elaborate approach, which concludes the application of Nelder-Mead solver point-wisely on the each of the controlling windows, with an aim of the best $I\left(t\right)_{\tau}^{\tau+5\Delta t}$ combination prediction. 

Whilst PID regulation works as fine as expected, point-wise application of the Nelder-Mead Fig. \ref{fig:OptimRes}c) solver on test slice windows results are rather unsatisfactory, especially in terms of runtime (Tab. \ref{table:1}). This is also the case for other commonly used solvers, as it shown in Appendix. \ref{sec:analysis:solvers}. Possible, runtime may be improved with a proper search for solvers parameters, which overall results in the grid-search with coordinate descent.

\begin{table}[!h]
    \centering
    
    \begin{tabular}{||l|l|l|l||}
        \hline
        \hline
        & \textbf{Autoreg ML} & \textbf{PID} & \textbf{Nelder-Mead}  \\
        \hline
        RMSE &     $\mathbf{1.15}$ &   $1.20$ &  $3.27$  \\
        \hline
        Optimization time, s &   $1.46$ &  $\mathbf{0.05}$ & $356.80$  \\
        \hline
        \hline
    \end{tabular}
    \caption{Quantitative optimization results}
    \label{table:1}
\end{table}

The key advantage of the ML-driven optimization over PID controller is its scaling capabilities, as the $M_{\mathrm{optim}}\left(x, w\right)$ may be easily increased in size, to capture more complicated trends and dependencies. 

\begingroup
\setlength{\xfigwd}{2\columnwidth}
\begin{figure*}
    \centering
    \includegraphics[width=1.05\xfigwd]{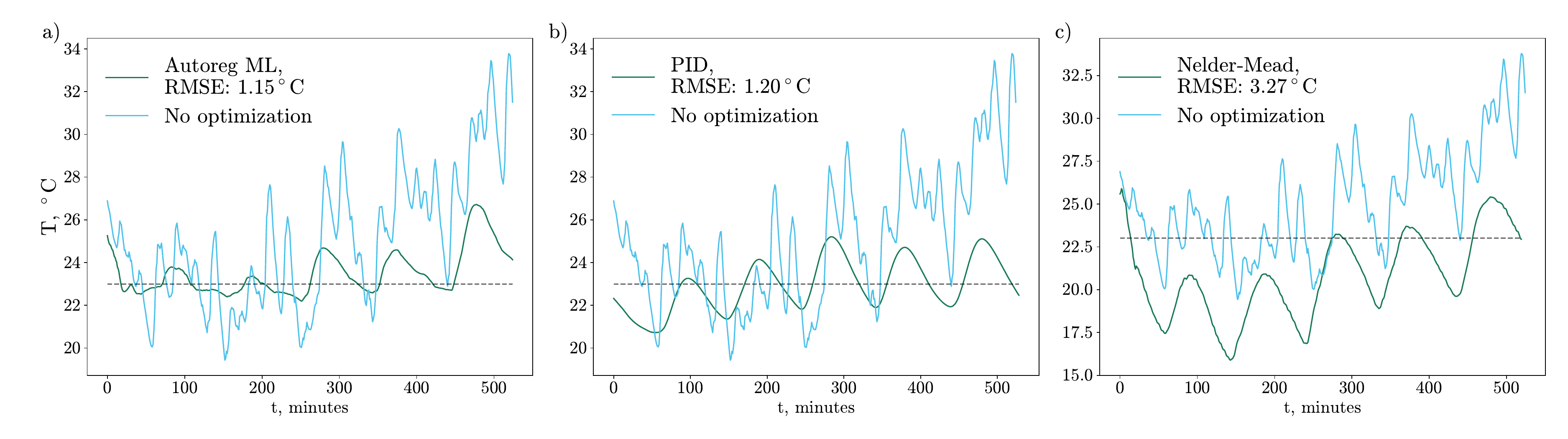}
    \caption{a) The optimization results obtained with ML-driven optimization, as Alg. \ref{algorithm:ML-driven-training} suggests, applied autoregressively. b) PID regulation results. c) Point-wise application of Nelder-Mead in terms of Alg. \ref{algorithm:classic-pointwise} statement}
    \label{fig:OptimRes}
\end{figure*}
\endgroup

\section*{Conclusion}\label{Conclusion}

In this paper, we present a possible approach of ML-driven optimization to industrial process stabilization. The proposed methodology revolves around the idea of substituting the best target values optimization problem with the weight optimization problem, suggesting the use of corresponding neural network to establish best optimization policy from the training on the response on the historical data. For that purpose, second neural network serving as the Oracle function, which replaces dynamical physical model, is utilized. 

The substitution shows its advantages over the classical point-wise approach, as it achieves the better RMSE (about $3$ times) of the resulting stabilization in significantly shorter time (sum of both training and optimizer application time) in the problem of room temperature control on previously unseen part of data. 

This approach is possible because the proposed predictor model captures the physical properties of the given industrial problem with amazing quality in terms of MSE ($0.002 \div 0.01$), thus not limiting the optimizer capabilities and its scalability to even more complicated problems.

Comparing the proposed technique with the classical PID control over a strictly specified physical model, it can be seen that the ML-driven optimization gives slightly better results ($\approx 10\%$), while requiring significantly more computation time.

The main advantage of the proposed methodology is that it can be easily applied to any more sophisticated industrial problem with higher number of dimensions in both time and number of parameters scales.  

\bibliography{lib}
\bibliographystyle{unsrt}

\appendix

\section{Point-Wise Solvers Application Review}\label{sec:analysis:solvers}
To give a larger perspective on comparison of point-wise optimization to proposed ML-driven one, other classical solvers repeatedly used in industry \cite{Ni_Brockmann_Darbandi_Kriegel_2025} evaluated and compared against our methodology. 

\begin{table}[!h]
    \centering
    \begin{tabular}{||l|l|l|l||}
        \hline
        \hline
        & \textbf{Dual Annealing} & \textbf{Powell} & \textbf{Cobyla}  \\
        \hline
        RMSE &     $\mathbf{2.77}$ &   $10.70$ & $3.00$  \\
        \hline
        Optimization time, s &   $1904.33$ &  $\mathbf{31.83}$  &  $56.69$\\
        \hline
        \hline
    \end{tabular}
    \caption{Quantitative optimization results for commonly used point-wise solvers in Scipy}
    \label{table:3}
\end{table}

\end{document}